\documentclass[a4paper,fleqn]{cas-dc}

\usepackage[numbers]{natbib}

\begin{document}
\let\WriteBookmarks\relax
\def\floatpagepagefraction{1}
\def\textpagefraction{.001}

\shorttitle{Plant Orbiter}

\shortauthors{Hartley et al.}

\title [mode = title]{Plant Orbiter: 3D Plant Reconstruction and Generation with Video Diffusion Orbit Cameras and 3D Gaussian Splatting}

\author[1]{Zane K.J. Hartley}[auid=000, bioid=1, orcid=0000-0002-1523-8885]
\fnmark[1]
\ead{zane.hartley@nottingham.ac.uk}
\credit{Conceptualization of this study, Methodology, Software}

\author[1]{Yong En Kok}[orcid=0000-0002-5620-9263]
\ead{Janet.Kok2@nottingham.ac.uk}
\fnmark[1]

\author[1]{Lewis A. G. Stuart}[orcid=0009-0004-0908-4148]
\ead{Lewis.Stuart@nottingham.ac.uk}

\author[1,2]{Andrew P. French}[orcid=0000-0002-8313-2898]
\ead{andrew.p.french@nottingham.ac.uk}

\author[1]{Michael P. Pound}[orcid=0000-0002-5016-1078]
\cormark[1]
\ead{michael.pound@nottingham.ac.uk}

\affiliation[1]{organization={University of Nottingham},
    addressline={School of Computer Science, Jubilee Campus}, 
    city={Nottingham},
    postcode={NG8 1BB}, 
    country={United Kingdom}}

\affiliation[2]{organization={University of Nottingham},
    addressline={School of Biosciences, Sutton Bonnington Campus}, 
    city={Loughborough},
    postcode={LE12 5RD}, 
    country={United Kingdom}}
    
\cortext[cor1]{Corresponding author}

\fntext[fn1]{Equal Contribution.}

\begin{abstract}
Modern digital phenotyping pipelines require the precise measurement of a range of 3D morphological plant traits. The accurate, high throughput acquisition of these traits helps breeders and researchers accelerate the development of new varieties, techniques and products. Recent advances in 3D technologies have enabled a rapid transition from 2D to 3D pipelines, but high throughput acquisition of digital 3D twins of plants remains a significant challenge due to complex morphology, thin structures and occlusions. 

In this paper we present Plant Orbiter, a new approach to generating 3D plants that leverages recent advances in video diffusion to produce high quality 3D Gaussian Splats of plants from a single image. We show that our method more accurately recreates plant morphology than other state-of-the-art methods, achieving the highest 3D F-score against all baselines, while being more practical for high throughput pipelines due to easier alignment and reduced cost of camera placement. Combined with a 2D diffusion model, we show that our method also achieves state-of-the-art results in the generation of 3D plant models from a text prompt.
\end{abstract}


\maketitle

\section{Introduction}
\label{sec:intro}

Digital 3D phenotyping presents substantial advantages in crop understanding over existing 2D analysis tools. Non-invasive methods for extracting accurate morphology enables a deep understanding of plant health~\cite{he2025multimodal}, and can accelerate the development of new varieties, as well as providing better monitoring tools for growers~\cite{hartley2023unlocking}. 3D plant analysis unlocks many of the most important measures for growers, such as leaf area, and fruit volume while handling occlusion for dense canopies better than can be achieved with a single view. Within mainstream computer vision, the rapid development of 3D reconstruction~\cite{mildenhall2021nerf, kerbl2023gaussians} and generation~\cite{xiang2025structured, zhao2025hunyuan3d} techniques has already lead to widespread adoption of 3D models for both generation and analysis. 

Despite this, no current approach on plants achieves sufficient morphological accuracy, or the rapid acquisition, that would be required for the high throughput phenotyping pipelines at scale needed for agricultural settings. Existing methods in 3D plant capture often exhibit one of a range of limiting factors. Hardware dependant approaches such as laser scanners or robotic capture environments have high upfront costs, and are often low throughput. Multi-view methods like Neural Radiance Fields (NeRF)~\cite{mildenhall2021nerf} or 3D Gaussian Splatting (3DGS)~\cite{kerbl2023gaussians} directly learn a 3D representation from image sequences around a plant, but rely on computationally expensive camera recovery from structure-from-motion, which has high failure rates on complex biological structures. To tackle this, single image or monocular reconstruction methods have emerged in recent years to avoid the need for complex calibration of camera parameters. Image-to-3D generators~\cite{hunyuan3d2025hunyuan3d, xiang2026native} attempt to learn a direct mapping from a single image to a 3D model, but often suffer from poor performance on plants, and their output is challenging to realign and scale to the input image~\cite{gao2026evaluation}. Novel view diffusion methods~\cite{chen2024v3d, voleti2024sv3d}, such as one included in our own pipeline, have not previously been applied to challenging domains like plants.

To improve the performance of reconstruction and other digital phenotyping tasks, the use of synthetic training data has proven increasingly popular due to a lack of available training datasets~\cite{ubbens2018use,hartley2024domain}. 3D data is often more challenging to produce and label, so this scarcity is even more pronounced and has motivated recent work in the generation of synthetic 3D plants using procedural and generative models~\cite{hartley2025plantdreamer} alongside wider research in text-to-3D models~\cite{zhao2025hunyuan3d, yi2024gaussiandreamer}. Despite recent advances, these methods suffer from high upfront costs, the need for hand-built per species rules, and in the case of generic models, poor performance on highly specific plant domains.

In this paper we present Plant Orbiter, a versatile pipeline for 3D reconstruction of accurate, automatically aligned plants. We leverage recent advances in video diffusion models to produce an accurate, geometrically consistent orbital video around the plant based on a single input image, from which 3D Gaussian splatting can produce an accurate model. Our method both reduces the practical and computational costs compared to multi-image capture setups, and recreates plants with higher geometric accuracy compared to previous camera guidance methods ~\cite{zhou2025stable, voleti2024sv3d}.

By prepending a text-to-image diffusion model to our pipeline, we also create a best in class model for the generation of entirely synthetic 3D plants that accurately adhere to a text prompt. We evaluate both the quality of synthetic plants but also the ability to prompt for specific phenotypic measurements, such as the number of fruits, and show that Plant Orbiter outperforms other leading text-to-3D models for both shape and prompt adherence.

The contributions of this paper can be summarised as follows:
\begin{itemize}
    \item A single image 3D reconstruction pipeline that achieves state-of-the-art geometric accuracy on reconstruction of plants.
    \item Experimental results for highly accurate height estimation from single image reconstruction, showing that video diffusion methods using our pipeline outperform 3D generation methods.
    \item A synthetic data generation pipeline that produces realistic synthetic plants, as well as experimental results showing our pipeline scores the highest for prompt adherence for phenotypical measurements.

\end{itemize}

\begin{figure*}[t]
  \centering
  \includegraphics[width=\textwidth]{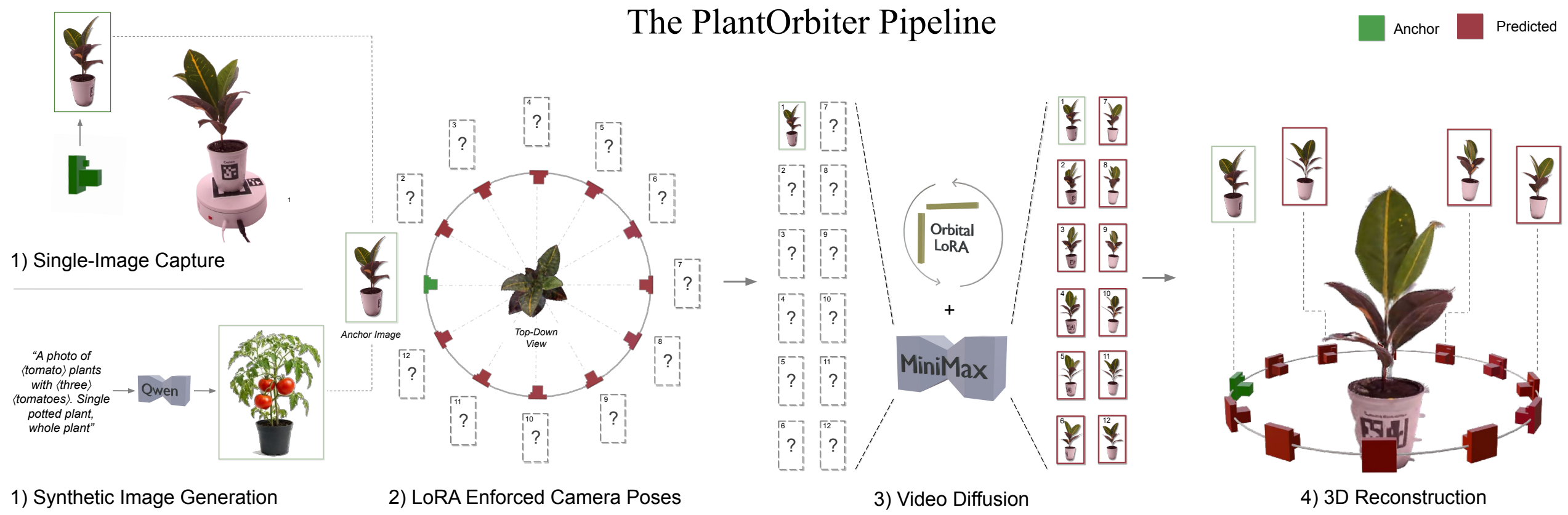}
  \caption{The plant orbiter pipeline, showing both the reconstruction and text-to-3D setups. In step 1, we acquire an appropriate input image of a plant, either a single photograph or the output of a high-quality 2D diffusion model. Step 2 and 3 show how a pretrained LoRA model is used in conjunction with MiniMax H3 to generate a 73 (figure shows 12 for simplicity) frame video with each frame at a predetermined position, forming an orbit around the photographed plant. Finally in step 4 we reconstruct the plant from the video using 3D Gaussian splatting, using the knowledge of the camera placements to both determine camera positions during initialisation, and to align the output plant to the input images frame.}
  \label{fig:main_pipeline}
\end{figure*}

\section{Related Work}
\label{sec:related work}

\noindent\textbf{3D plant reconstruction.}
3D plant reconstruction makes it possible to capture detailed, precise plant morphology traits and supports automated, high-throughput crop phenotyping~\cite{li2025survey}. 3D plants models have traditionally been obtained with laser scanning~\cite{lin2015lidar}, structured light or RGB-D sensors~\cite{choi2024nerf}, or from many overlapping photographs using structure-from-motion and multi-view stereo~\cite{wang2025accurate}. These approaches can be highly accurate, but they need dedicated equipment, controlled conditions and many viewpoints per plant. Plant structure also makes them fragile; thin leaves, overlapping organs and self-occlusion result in gaps and distortions in the reconstruction~\cite{li2025survey,okura20223d}.

Neural radiance fields (NeRF)~\cite{mildenhall2021nerf} and 3D Gaussian splatting (3DGS)~\cite{kerbl2023gaussians} have since substantially raised the fidelity of plant reconstruction from ordinary photographs. While NeRF variants successfully model complex appearance without depth supervision, their implicit representation makes it hard to extract the explicit surfaces needed for trait measurement~\cite{arshad2024evaluating,shen2025plantgaussian}. 3DGS instead models the scene with explicit Gaussian primitives, which train quickly and render in real time, recovering traits such as plant height and leaf dimensions with accuracy close to laser scanning ~\cite{stuart2025high,ma2026plant3r,yang2025gaussianplant}. Robotic capture systems extend them to occluded organs and growth tracking ~\cite{adebola2025botany, adebola2025growsplat}. However, these gains in fidelity have not reduced the cost of acquisition, as current methods still require tens to hundreds of calibrated photographs of each plant.%

\noindent\textbf{3D generation from images and text.}
Generative 3D models learn shape and appearance from 3D training data that let them produce a complete object, including its unseen sides, from a single image or a text prompt. Because 3D training datasets are far scarcer than image-text pairs, early text-to-3D methods optimised a 3D representation for each prompt by distilling a pretrained 2D diffusion model through score distillation sampling (SDS)~\cite{poole2022dreamfusion}. GaussianDreamer~\cite{yi2024gaussiandreamer} applies SDS to 3D Gaussians for speed. MVDream~\cite{shi2024mvdream} distils from a camera-conditioned multi-view diffusion model, which reduces the multi-face artefacts that arise when each view is supervised independently. In parallel, image-conditioned methods learned to synthesise novel views from one image~\cite{liu2023zero, shi2023zero123++, liu2024syncdreamer, long2024wonder3d} and then lifted them to 3D in a single forward pass~\cite{hong2024lrm, tang2024lgm, xu2024instantmesh}. TRELLIS~\cite{xiang2025structured} decodes structured sparse-voxel latents into Gaussians or meshes from an image or a prompt, and TRELLIS.2~\cite{xiang2026native} extends this to open and non-manifold surfaces. Hunyuan3D 2.1~\cite{hunyuan3d2025hunyuan3d} first generates the shape with a diffusion transformer and then textures it in a separate stage, and DiffSplat~\cite{lin2025diffsplat} repurposes an image diffusion model to generate Gaussian splats.

Multi-view image models generate only a handful of views, whereas video diffusion models, trained on vast real-world footage, learn how objects transform under continuous camera motion and can synthesise dense novel-view trajectories~\cite{blattmann2023stable}.
SV3D~\cite{voleti2024sv3d} and V3D~\cite{chen2024v3d} fine-tune Stable Video Diffusion on rendered 360\textdegree{} orbits of synthetic objects, so that a single image yields a full orbit at known camera angles, to which a NeRF or Gaussian splat is then fitted. Later work improves the resolution and consistency of these orbits~\cite{yang2024hi3d, zuo2024videomv,melas20243d, wang2026towards}, and SEVA~\cite{zhou2025stable} extends them to arbitrary camera trajectories from one or more input views.
Camera control can also be added to general-purpose video generators, either by injecting camera poses~\cite{wang2024motionctrl, he2025cameractrl} or by training lightweight LoRAs for specific camera motions~\cite{sun2025dimensionx}.
VIST3A~\cite{go2025vist3a} skips explicit frames and decodes scene-level 3D Gaussians directly from the latents of a text-to-video model.

These models are trained mainly on synthetic 3D assets~\cite{deitke2023objaverse} and evaluated on general objects, and plants, with their thin, overlapping and highly variable organs, expose the limits of this training~\cite{hartley2025plantdreamer, gao2026evaluation}. Single-image methods built for plants do exist, but they mostly target trees and rely on hand-designed botanical priors~\cite{li2021trees, li2024svdtree, lee2024treedfusion, liu2025hierarchical}. From text, general 3D generators also offer little control over specific plant traits, such as the number of fruit, whereas text-to-image models, trained on far more data, follow such detailed prompts much more closely~\cite{zhao2026qwen}. Our aim is therefore to generate 3D plants that are faithful to an input image or text prompt, leveraging the priors of existing text-to-image and video diffusion models without requiring specialised 3D plant training data.

\section{Method}
\label{sec:method}

\subsection{Single Image Orbital Video}
\label{subsec:orbitvideo}
Our method leverages the strong understanding of 3D shape that has been learned by powerful video diffusion models. We use Minimax H3~\cite{minimax2026h3} along with a LoRA trained to produce orbital camera videos around the plant ~\cite{dawson2026orbitlora}. For each video the model is prompted with a single image of the front side of a given plant. We preprocess each image to remove all background elements using provided masks or SAM 2.1~\cite{ravi2025sam} where masks are not provided. We then undistort the image, add it to a white canvas, and resize it to fit Minimax's native $768 \times 768$ resolution. The input image is given the Rotary Position Embedding (RoPE)~\cite{su2024roformer} position for the first and last frame in the sequence, ensuring the camera arc returns perfectly to the original location in the loop. Additionally, the model receives the text prompt used to train the original LoRA, conditioning the model to produce a consistent, level camera arc around the object while keeping the plant still to discourage motion that would otherwise reduce the model quality and enforce a static scene.

\subsection{Known Pose Camera Recovery}
\label{subsec:knownposes}
Generally, the first stage of 3DGS pipelines is the extraction of camera positions using COLMAP~\cite{schonberger2016structure}, using structure from motion to correctly identify the cameras position and rotation within the scene. For 3D plant reconstruction, this method struggles to extract features on the repetitive, detailed structures of the plant surfaces, as well as being computationally expensive. Instead we exploit the consistent, known parameters of the orbital video to skip this stage of reconstruction, and use easily calculated camera poses based on a known camera path. With 73 frames including the duplicated frame for 0 and 72, each frame represents a uniformly spaced 5 degree increment in the orbit around the plant starting from an initial pose 1 arbitrary unit from the rotational axis. The camera's path maintains a fixed distance and directed at this centre axis, which the video diffusion model takes to be the object at the horizontal midpoint of the frame. 

For effective training, 3D Gaussians are conventionally initialised from an initial point cloud generated from the structure-from-motion process employed to extract the camera poses. However, as COLMAP is skipped, we replace its initial Gaussians with ones we compute from a visual hull of the plant, ensuring high-quality 3D reconstruction. Each video frame is used to carve a silhouette cone~\cite{laurentini1994visual} from a voxel grid, with each surviving voxel used as an initial Gaussian~\cite{yang2024gaussianobject, chen2024v3d}. As the video model does not output its focal length or elevation, we also calculate an estimate of these values during this process by determining the values that cause the projected carved hull to intersect most accurately with the plant masks across all viewpoints.

\subsection{3D Reconstruction}
\label{subsec:3dreconstruction}
Training each Gaussian splat is performed using NerfStudio's~\cite{tancik2023nerfstudio} Splatfacto~\cite{kerbl2023gaussians, ye2025gsplat} model with default parameters, apart from enabling antialiased rasterisation and scale regularisation (with maximum Gaussian axis ratio of 3) to remove long, needle like Gaussians. We perform masked training so that only the plant is reconstructed, and the 3D model is trained for 30,000 iterations, which is standard practice for 3DGS reconstruction. For each plant, we also produce a matching 2D Gaussian Splat (2DGS)~\cite{huang20242d} model, which replaces the 3D gaussians with flat 2D disks and is trained with additional depth-normal consistency and depth distortion losses. Because of this 2DGS models produce more surface-aligned reconstructions compared to the original 3DGS implementation, allowing for a mesh to be extracted with truncated signed distance function (TSDF)~\cite{curless1996volumetric} after training which we then use for our 3D geometric evaluation. 

\subsection{Alignment}
\label{subsec:alignment}
In order to accurately evaluate the 3D plant reconstructions, alignment between generated models and the ground truth must first be established. All 3D reconstruction models return the plant within their own arbitrary canonical frame, which require a transform (scale, rotation, and translation) to bring them into the frame of the real cameras for evaluation. This reflects real world usage where the ability to accurately phenotype from a reconstruction relies on it being correctly scaled and positioned within its frame; however precise realignment is a known challenge~\cite{gao2026evaluation}.

To precisely position reconstructed plants within the real scene, a known real world measure is required to determine the scale of the reconstructed object. In a capture environment designed with this in mind, an object of known dimensions can be used as a reference. As a reference is not provided in each dataset we use, we instead calculate the real world scale of each plant from the plant's distance from the input camera, which we derived from each dataset's calibrated camera pose data. We then scale each reconstructed object such that the orbit radius matches this distance, adjusting for different focal lengths. After scaling, each reconstruction must be orientated and positioned to match the position of the ground truth. For Plant Orbiter, we initialise this process by aligning the first orbit camera to the input image's camera. Afterwards, we fit the reconstruction's silhouette to the plant mask as seen from the input image (or for oracle placement, all held out view) over 300 steps using an Adam Optimiser. This gives a similarity correction using uniform scale, 3D rotation and translation which provides our final alignment. 

For the 3D generator baselines evaluated against in this paper we cannot rely on a known start position or radius, since these models produce a 3D reconstruction in an unknown canonical frame depending on the plant and generator. Instead we perform scaling by matching the reconstructed plant's height against the pixel height in the input image. Then to normalise each reconstruction, we orient the plant upright using the generators default up axis, or using its longest axis with its pot as the base where this method fails. We then iterate over 10\textdegree{} yaw increments over 0.8, 1.0 and 1.25 scales and rank for silhouette IoU against the alignment view(s). We double the number of steps taken during correction to maximise the precision of the final alignment, ensuring a fair comparison between plant reconstructions.

To match a real world scenario where only a single image is captured, we compute this similarity against only the single image used to generate the 3D model (experiments labelled \textit{input view placement}), however we also complete a second evaluation round aligning against all held out views (experiments labelled \textit{oracle}), to determine true accuracy score assuming an upper bound alignment.

\subsection{Text-to-3D generation}
\label{subsec:textto3d}
Since Plant Orbiter needs only a single image, it can be turned into a text-to-3D generator by first generating a synthetic image from a text prompt using a text-to-image model, and then lifting it to 3D with the pipeline described earlier. This approach inherits the strong prompt adherence of standard 2D text-to-image models, which are trained on far more data than is available in 3D datasets, allowing fine-grained attributes such as the number of fruit to be specified in the prompt. 

We use Qwen-Image-2.1~\cite{qwen2026qwenimage21} with default parameters to generate $2048 \times 2048$ pixel input images. To ensure the generated image matches the input expected by the orbital video model, we append a fixed framing suffix to the content prompt. For example, the prompt for bean becomes ``A photo of bean plants. Single potted plant, whole plant and pot fully in frame and centred with margin around it, eye-level side view, isolated on a plain white background, studio lighting.'' The suffix is identical for every plant species generated, and adds only framing, background and lighting terms, so the plant description itself is the same prompt that every baseline receives.

\section{Single-Image Reconstruction Experiments}
\label{sec:img_recons_experiments}

\subsection{Datasets}
\label{subsec:datasets}
For our experiments, we evaluate over a range of different multi-view plant datasets captured using a combination of cameras mounted on robot arms, turntables, bespoke multi-camera capture environments and handheld capture. For each plant we select one real photo showing the entire plant, which is then used as the input for all single view reconstruction. 

The PlantDreamer~\cite{hartley2025plantdreamer} dataset constitutes the primary dataset we evaluated, containing multi-view captures of bean, mint and kale plants captured using two Universal Robots UR5 robotic arms. The overall dataset contains 21 individual plants, including 12 instances of bean plants, 6 instances of mint and 3 instances of kale. For our experiments, we omit a single instance of bean due to incorrect mask data included in the published dataset. Each plant is captured with approximately 220 views at a range of elevations, and 27 images were included in the evaluation dataset for assessing the photometric quality of the reconstruction. For a fair comparison, these evaluation images were solely used as the reference images in the single-image generation models. PlantDreamer does not come with a true 3D ground truth and instead provides point clouds derived from 3D Gaussian splats \cite{A_G_Stuart_2025_ICCV}, which are produced using the images provided which we use as a 3D reference during experimentation. During experimentation we observed that the masks provided with PlantDreamer were poorly cropped to the plant, resulting in poor performence, so applied an intersection between the provided masks and IS-Net~\cite{qin2022highly} for all experiments to improve the segmentation.

The Berkeley Botany-Bot~\cite{adebola2025botany} dataset contains 8 different houseplants and was captured by placing the plant on a turntable, acquiring images at four elevations. Each plant has 91-96 total views, and we evaluate using all held-out samples for each experiment. Botany-Bot neither provides a 3D ground truth nor an appropriate 3D reference, so instead we generate our own 2DGS from the real photos provided as a 3D reference.

The PLANesT-3D~\cite{mertoglu2024planest} dataset contains 10 instances of potted pepper and rose plants, each captured manually using a handheld DSLR camera. These plants were captured from 180-400 viewpoints and evaluated using 64 evenly spaced evaluation views taken from the unused images. We exclude a single pepper instance during 3D evaluation because its point cloud data could not be aligned to its photos, suggesting movement during capture. For 3D, PLANesT provides labelled Metashape point clouds which are registered to the cameras by a silhouette fit.

The GrowSplat~\cite{adebola2025growsplat} dataset, by contrast to the previous methods, is a time series dataset showing a quinoa and a sequoia plant each growing over 17 time steps. For the purposes of our experiments, we remove the earliest time step for quinoa, as the plant at that stage was too small to be reconstructed by any of the methods evaluated in this paper. The images were captured using an imaging box that contained fixed cameras at five positions around the plant, at three fixed heights, resulting in a total of 15 views per plant. We evaluate over all 14 unused held out images during our evaluations. Each plant in GrowSplat also provides a metric plant point cloud using an accurate LiDAR scanner, which we use as a height reference for our downstream evaluations.

For all datasets the background was digitally removed, using either provided masks to isolate the plant and pot from the background, or our own masks generated with a SAM 2.1~\cite{ravi2025sam} segmentation pipeline. This was done to cleanly produce a Gaussian splat of the target plant and remove background artifacts in the orbit video that did not consistently rotate with the plant. 

\subsection{Baselines}
\label{subsec:Baselines}
To evaluate our method we select a number of SotA monocular 3D reconstruction methods to compare against. We include TRELLIS~\cite{xiang2025structured}, TRELLIS.2~\cite{xiang2026native}, and Hunyuan3D 2.1~\cite{hunyuan3d2025hunyuan3d} (the most recent version available for local use), as these are currently the highest performing available models that could be employed as part of a 3D phenotyping pipeline, and have seen interest in previous phenotyping works for single-image reconstruction ~\cite{gao2026evaluation}. TRELLIS produces its own Gaussians and a mesh for 3D scoring, while both TRELLIS.2 and Hunyuan3D 2.1 produce textured meshes which we convert into Gaussians for rendering, and are run using default settings as described in their documentation. Each method receives the same input image having undergone the same preprocessing steps as Plant Orbiter. We opt to give each baseline our own cropped image, rather than relying on built in background removal, as we found poorer performance at this stage unfairly penalised these models relative to our own. 

We additionally test against SEVA~\cite{zhou2025stable} and SV3D~\cite{voleti2024sv3d}, two other methods that use video diffusion for 3D reconstruction. SEVA is a multi-view diffusion model that we configure to generate 72 views around the object, matching our setup, while SV3D is a video diffusion pipeline, from which we configured to generate 21 frames around the plant at 10\textdegree{} of elevation. As neither of these models natively generate a 3D reconstruction, we append \textit{our} downstream pipeline, including use of known poses, as well as splatting and meshing pipelines to produce the outputs used in our evaluation. As such, this test primarily evaluates the accuracy of the orbital LoRA compared to standard video diffusion approaches.

To fairly evaluate the baseline outputs, proper alignment between the 3D plant models was needed. Reconstruction models generally produce output in their own canonical frame, and reorienting and scaling them is a known challenge~\cite{gao2026evaluation}. A core advantage of video diffusion method is the implicit alignment with the input image that is automatically achieved, while baseline methods were aligned by matching the reconstructions silhouette against the 2D mask. Full details on alignment are found in section \ref{subsec:alignment}. Finally, to improve reliability of our results, all experiments on reconstruction quality in sections \ref{subsec:singleimagereconstruction} and \ref{subsec:3Dgeometry} were carried out using three different seeds.

\subsection{Evaluation Metrics}
\label{subsec:reconstructionmetrics}
We perform evaluation across three sets of experiments shown in tables \ref{tab:main2d}, \ref{tab:geometry}, and \ref{tab:height}, using metrics defined below.

\textit{2D evaluation} is performed using a range of metrics which directly compare renders of the reconstruction against ground truth held out images of the same camera pose. We use four metrics: IoU (the overlap between the render and ground truth silhouettes divided by their union), PSNR (the pixelwise error between the render and ground truth), LPIPS~\cite{zhang2018unreasonable} (deep feature distance between the render and ground truth image), and DISTS~\cite{ding2020image} (comparison of structure and texture between render and ground truth, included for being more tolerant to small misalignments). For fairness, especially with pixel error metrics, we apply colour correction to the renders of each method's output using a transform fitted to each input image.

While 2D IoU already gives some indication of 3D structural accuracy, we also include \textit{3D evaluation} where we compare the aligned 3D reconstruction against a ground truth or derived 3D representation from each full dataset. For this we use Chamfer Distance (the average distance between each point on the reconstruction to the nearest point on the ground truth, and again from the ground truth to the reconstruction both averaged, normalised to the reference bounding box and shown as $\times10^{3}$) and F-score (the harmonic mean of the fraction of points in the reconstruction within $\tau$ of the reference and vice versa, where $\tau$ is 1\% of the reference bounding box diagonal).

\subsection{Single Image Reconstruction}
\label{subsec:singleimagereconstruction}

\begin{figure*}
  \centering
  \includegraphics[width=\textwidth]{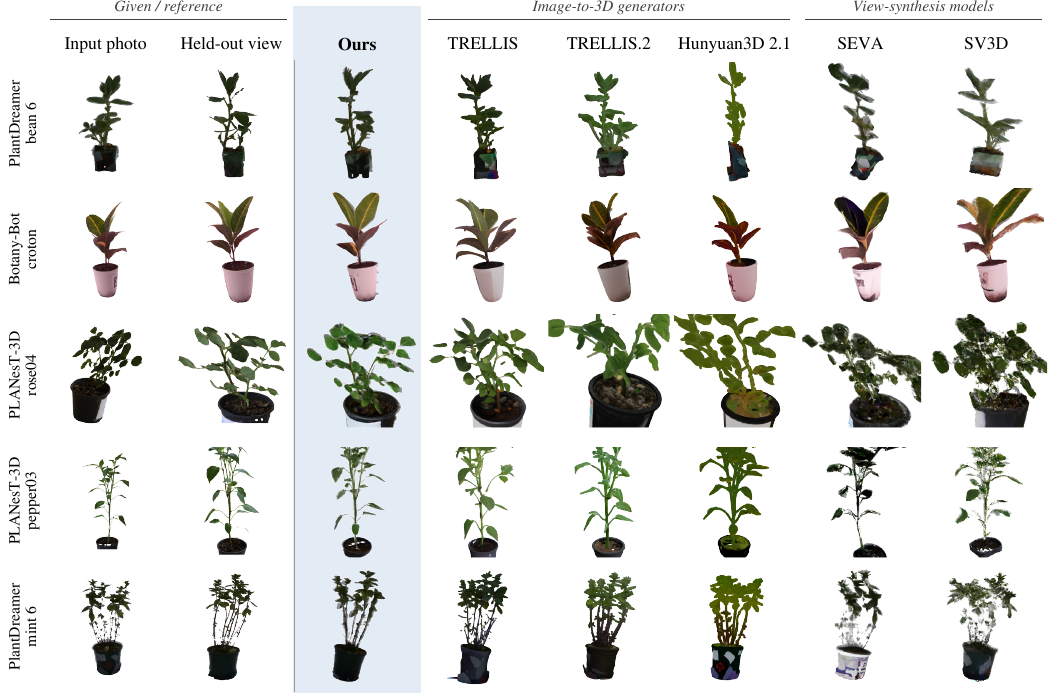}
  \caption{Outputs of our model and all baselines showing examples from each test set. Column 1 shows the input image used for single image 3D reconstruction and column 2 shows the held-out image from an unseen view. For each output we render the reconstruction from the held-out view using oracle placement, demonstrating each objects ability to accurately recreate the occluded side and plant geometry from the front image.}
  \label{fig:singlephotoreconstruction}
\end{figure*}

Results in Table \ref{tab:main2d} demonstrate that for the input-view case, we score best for shape measures for PlantDreamer (0.552 vs 0.529) and Botany-Bot (0.556 vs 0.504) and score second on the PLANesT (0.343 vs 0.348). For DISTS measurements of perceptual similarity, we score best across all datasets, while for pixel metrics PSNR and LPIPS we are among the top three best performing models. Visualisations of all models results for all datasets can be seen in Fig.~\ref{fig:singlephotoreconstruction}.

Table \ref{tab:main2d} shows separate results for input-view placement and oracle placement. These results show substantial performance improvements are gained by all models through more accurate placement, and notably that the improvement of the TRELLIS and Hunyuan3D models was larger than those of our method, with TRELLIS.2 now leading in some metrics having received the largest improvement due to alignment. By contrast, Plant Orbiter continues to lead Botany-Bot across all four metrics, as well leading in IoU for PLANesT. This shows that placement from a single frame is unreliable for the generator baselines, whereas our method is much stronger in input-view placement due to orientation being known by construction. We find that the oracle placements for the generator models have median 16\textdegree{} (TRELLIS.2), 27\textdegree{} (TRELLIS) and 48\textdegree{} (Hunyuan3D) difference between input placement and oracle placement, demonstrating that single image alignment is repeatedly insufficient to align the outputs of these models. 

The input-view alignment best reflects the real world situation where an agronomist would want to take measurements using a single hand-captured image, and it is in this setting that our method performs best. We also observe that even with Oracle placement, our method still achieved the second highest score for PlantDreamer, and the highest score for Botany-Bot and PLANesT, despite the latter being the most challenging dataset. This shows that our methods advantage is not from alignment alone, but also matches or exceeds the strongest generator on shape, and we argue geometric measures are the most important for downstream phenotyping which is supported further in section \ref{subsec:3Dgeometry}.

We do observe that while our method performs best in spatial and perceptual metrics (IoU and DISTS respectively), our method is less comparatively strong in pixel similarity metrics such as PSNR, where SV3D is the strongest overall method. Characteristics of pixel similarity metrics mean that fine details with small misalignments are punished more, while lower detail, and blurred geometry is more highly rewarded. We also observe that the absolute PSNR values for all methods are generally poor, likely caused by the invented details on the occluded side of the plant. We believe for these reasons PSNR is less reliable as a signal for the overall quality of reconstruction for these datasets. 

Overall, our results show that our method is the best performing model on shape and perceptual accuracy, with substantial gains over the currently favoured generator models, such as TRELLIS.2, for the real world case of single image generation and alignment. We also show that our method creates the geometrically highest performing reconstructions on most datasets, even in the ideal case of oracle alignment.

\begin{table*}
  \caption{Monocular Reconstruction scored in 2D on unseen images. Results shown for input-view placement, where alignment is performed using the input image only and oracle-view placement, where all held out images are used. We evaluate 2D geometry (IoU), pixel accuracy (PSNR) and feature alignment (LPIPS and DISTS). Best reconstruction results for each dataset and metric are shown in bold, second best are shown underlined. PlantDreamer and Botany-Bot are abbreviated to PD and BB respectively. Full capture indicates scores for splats reconstructed from all non-test real images and acts as an upper bound.}
  \label{tab:main2d}
  \centering
  \footnotesize
  \setlength{\tabcolsep}{4pt}
  \begin{tabular}{@{}lccc@{\hspace{8pt}}ccc@{\hspace{8pt}}ccc@{\hspace{8pt}}ccc@{}}
    \toprule
     & \multicolumn{3}{c}{IoU$\uparrow$} & \multicolumn{3}{c}{DISTS$\downarrow$} & \multicolumn{3}{c}{PSNR$\uparrow$} & \multicolumn{3}{c}{LPIPS$\downarrow$} \\
    \cmidrule(lr){2-4} \cmidrule(lr){5-7} \cmidrule(lr){8-10} \cmidrule(lr){11-13}
    Method & PD & BB & PLANesT & PD & BB & PLANesT & PD & BB & PLANesT & PD & BB & PLANesT \\
    \midrule
    & \multicolumn{12}{c}{\textit{Input-view placement}} \\
    \cmidrule(l){2-13}
    TRELLIS & 0.406 & 0.446 & 0.269 & 0.225 & 0.229 & 0.246 & 5.75 & 7.86 & 5.24 & 0.083 & 0.089 & 0.099 \\
    TRELLIS.2 & 0.385 & \underline{0.504} & 0.281 & 0.227 & 0.220 & 0.238 & 5.44 & 8.56 & 5.62 & 0.084 & 0.085 & 0.095 \\
    Hunyuan3D 2.1 & 0.336 & 0.482 & 0.288 & 0.262 & 0.241 & 0.239 & 5.01 & 8.78 & \underline{5.98} & 0.089 & 0.087 & \underline{0.094} \\
    \addlinespace[3pt]
    SEVA & 0.448 & 0.442 & 0.285 & 0.215 & \underline{0.213} & 0.246 & 6.57 & 8.66 & 5.40 & \underline{0.082} & 0.083 & 0.097 \\
    SV3D & \underline{0.529} & 0.495 & \textbf{0.348} & \underline{0.201} & 0.217 & \underline{0.234} & \textbf{8.25} & \textbf{9.27} & \textbf{6.23} & \textbf{0.072} & \underline{0.081} & \textbf{0.093} \\
    \cmidrule(l){1-13}
    Ours & \textbf{0.552} & \textbf{0.556} & \underline{0.343} & \textbf{0.199} & \textbf{0.194} & \textbf{0.226} & \underline{6.75} & \underline{9.19} & 5.74 & \underline{0.082} & \textbf{0.080} & 0.095 \\
    \midrule
    & \multicolumn{12}{c}{\textit{Oracle placement}} \\
    \cmidrule(l){2-13}
    TRELLIS & 0.585 & 0.557 & 0.333 & 0.190 & 0.207 & 0.216 & 8.02 & 9.08 & 6.73 & 0.074 & 0.081 & 0.091 \\
    TRELLIS.2 & \textbf{0.620} & \underline{0.581} & 0.344 & \textbf{0.179} & 0.208 & \textbf{0.204} & \underline{9.02} & 9.63 & \underline{7.38} & \textbf{0.069} & 0.079 & \underline{0.087} \\
    Hunyuan3D 2.1 & 0.522 & 0.578 & 0.366 & 0.215 & 0.227 & 0.206 & 7.65 & 9.75 & \textbf{7.68} & 0.076 & 0.081 & \textbf{0.086} \\
    \addlinespace[3pt]
    SEVA & 0.547 & 0.543 & 0.363 & 0.197 & \underline{0.205} & 0.219 & 8.37 & 9.83 & 6.81 & \underline{0.073} & 0.079 & 0.091 \\
    SV3D & 0.591 & 0.575 & \underline{0.391} & 0.197 & 0.211 & 0.221 & \textbf{9.26} & \underline{9.87} & 7.18 & \textbf{0.069} & \underline{0.077} & 0.089 \\
    \cmidrule(l){1-13}
    Ours & \underline{0.612} & \textbf{0.643} & \textbf{0.408} & \underline{0.188} & \textbf{0.188} & \underline{0.205} & 8.21 & \textbf{10.19} & 6.87 & 0.074 & \textbf{0.074} & 0.090 \\
    \midrule
    \textit{Full capture} & 0.920 & 0.925 & 0.784 & 0.087 & 0.070 & 0.082 & 18.47 & 20.32 & 17.29 & 0.030 & 0.029 & 0.033 \\
    \bottomrule
  \end{tabular}
\end{table*}

\subsection{3D Geometry}
\label{subsec:3Dgeometry}
In order to evaluate our output 3D accuracy, we carry out an additional evaluation on Plant Orbiter against the baselines. Since the 3D Gaussian splats produced by our method have no explicit surface, we instead train a 2D Gaussian splat using the same generated orbit video as in our previous approach and extract a mesh from it, giving a surface that can be directly compared against the output from the other baselines. We evaluate over Chamfer distance and F-score to evaluate the true accuracy in 3D of our method and the baselines. Following alignment described in 3.4, we perform an additional normalisation step followed by 200 iterations of the rigid iterative closest point algorithm to properly align the 3D model with the reference point cloud. For PLANesT Dataset, we additionally crop the pot from the 3D model to match the contents of the reference point cloud.

We report all results in Table \ref{tab:geometry} with best performing methods highlighted in bold. The results show that our method outperforms all baselines in F-score and has the best Chamfer distance on all but PlantDreamer, where we score slightly worse than TRELLIS.2 (18.3 to 17.7). Within each dataset, we individually beat all baselines on a majority of individual plants, with the clearest wins on Botany-Bot and PLANesT. This highlights how our method excels at generating geometrically consistent plants between frames, which is important for accurate reconstruction and phenotyping.

\begin{table*}
  \caption{Monocular Reconstruction scored in 3D using reference point clouds. All reconstructed plants are aligned with oracle placement and further aligned with rigid ICP. For PLANesT we evaluate the plant only and digitally remove the pot for evaluation. We show results for geometric accuracy in 3D evaluated with Chamfer Distance and F-score. Best result for each dataset and metric are shown in bold, second best are shown underlined.}
  \label{tab:geometry}
  \centering
  \footnotesize
  \setlength{\tabcolsep}{4pt}
  \begin{tabular}{@{}lcccccccc@{}}
    \toprule
     & \multicolumn{2}{c}{PlantDreamer ($n{=}20$)} & \multicolumn{2}{c}{Botany-Bot ($n{=}8$)} & \multicolumn{2}{c}{Pepper ($n{=}9$)} & \multicolumn{2}{c}{Rose ($n{=}10$)} \\
    \cmidrule(lr){2-3} \cmidrule(lr){4-5} \cmidrule(lr){6-7} \cmidrule(l){8-9}
    Method & CD$\downarrow$ & F$\uparrow$ & CD$\downarrow$ & F$\uparrow$ & CD$\downarrow$ & F$\uparrow$ & CD$\downarrow$ & F$\uparrow$ \\
    \midrule
    TRELLIS       & 20.6 & 0.367 & 22.8 & 0.426 & 68.5 & 0.142 & 52.2 & 0.168 \\
    TRELLIS.2     & \textbf{17.7} & \underline{0.427} & \underline{22.6} & \underline{0.464} & 65.6 & 0.161 & 59.6 & 0.157 \\
    Hunyuan3D 2.1 & 34.8 & 0.300 & 24.4 & 0.424 & 44.5 & 0.208 & 51.0 & 0.204 \\
    \addlinespace[2pt]
    SEVA          & 30.8 & 0.332 & 28.5 & 0.378 & 39.5 & 0.216 & 42.0 & \underline{0.217} \\
    SV3D          & 20.9 & 0.410 & 23.5 & 0.434 & \underline{35.7} & \underline{0.230} & \underline{40.4} & 0.215 \\
    \midrule
    Ours          & \underline{18.3} & \textbf{0.443} & \textbf{15.9} & \textbf{0.540} & \textbf{30.2} & \textbf{0.252} & \textbf{37.1} & \textbf{0.257} \\
    \bottomrule
  \end{tabular}
\end{table*}

\subsection{Downstream Trait Evaluation}
\label{subsec:traits}
To evaluate our model for representing real downstream plant traits using 3D reconstruction, we select plant height estimation (a common type of spatial phenotyping task) that is well suited for evaluating 3D reconstruction approaches. For these experiments, we use the dataset GrowSplat, a temporal dataset showing two growing plants over a series of 17 time steps. By evaluating the ability of each reconstruction technique to measure the growing height of each plant over many time steps, we seek to verify firstly, to what extent 3D reconstruction methods can be used to measure major phenotypic traits such as height, and additionally to measure the performance of PlantOrbiter with respect to the baselines. We measure each plant from the base of the plant, as taken from the reference, removing the pot and soil prior to measuring. To remove any individual stray Gaussians above the top of the plant, we measure the height by downsampling to one point per 5mm voxel and take the 99.5th percentile of the points above its base.

\begin{figure}
  \centering
  \includegraphics[width=\columnwidth]{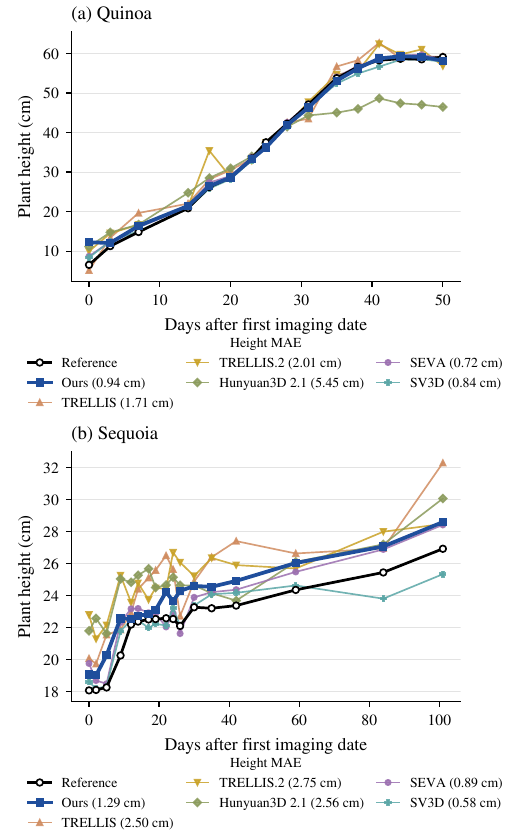}
  \caption{Plant height over time on GrowSplat, reconstructed from a single photo per imaging date.}
  \label{fig:fig_growth}
\end{figure}

We show numerical results in Table \ref{tab:height} and additional time series data in Fig.~\ref{fig:fig_growth}. The results show that the performance for downstream evaluation from all three \textit{camera-based} methods (mean MAPE 3.7 - 5.9\%) are much more closely able to estimate the plant heights compared to the generator methods (mean MAPE 12.7 - 15.8\%). While both SV3D and SEVA are the best models overall, our method is similarly much more performant in these tasks compared to Hunyuan3D 2.1, the current best generator model. For plant height in particular, our method closely tracks the recorded growth curve, with a slight 6\% overestimate of height, while in Fig.~\ref{fig:fig_growth} we see SEVA has less consistent tracking despite the lower average error. As all three of the best performing model use our reconstruction and camera placement models, we attribute their relatively much higher performance to our pipeline as opposed to the specific video models.

\begin{table*}
  \caption{Plant height estimation from one photo 3D reconstructions, as tested on GrowSplat (using 16 time steps for quinoa and 17 time steps for sequoia) and PLANesT pepper (n=9) and rose (n=10). We evaluate over Mean Absolute Percent Error (MAPE) and Spearman Rank Correlation $\rho$. For these experiments all reconstructions are aligned with input-view placement, to best represent single image performance in a downstream setting. Best result for each dataset and metric are shown in bold, second best are shown underlined.} 
  \label{tab:height}
  \centering
  \footnotesize
  \setlength{\tabcolsep}{4pt}
  \begin{tabular}{@{}lcc@{\hspace{8pt}}cc@{\hspace{8pt}}cc@{\hspace{8pt}}cc@{\hspace{8pt}}c@{}}
    \toprule
     & \multicolumn{4}{c}{GrowSplat (time series)} & \multicolumn{4}{c}{PLANesT-3D (hand-held)} & \\
    \cmidrule(lr){2-5} \cmidrule(lr){6-9}
     & \multicolumn{2}{c}{Quinoa} & \multicolumn{2}{c}{Sequoia} & \multicolumn{2}{c}{Pepper} & \multicolumn{2}{c}{Rose} & Mean \\
    \cmidrule(lr){2-3} \cmidrule(lr){4-5} \cmidrule(lr){6-7} \cmidrule(lr){8-9}
    Method & MAPE$\downarrow$ & $\rho\uparrow$ & MAPE$\downarrow$ & $\rho\uparrow$ & MAPE$\downarrow$ & $\rho\uparrow$ & MAPE$\downarrow$ & $\rho\uparrow$ & MAPE$\downarrow$ \\
    \midrule
    TRELLIS & 7.4 & \underline{0.98} & 11.2 & \textbf{0.95} & 28.9 & $-$0.13 & 12.7 & 0.90 & 15.1 \\
    TRELLIS.2 & 9.8 & 0.96 & 12.9 & 0.74 & 23.9 & $-$0.02 & 16.6 & 0.83 & 15.8 \\
    Hunyuan3D 2.1 & 17.0 & 0.97 & 12.0 & 0.57 & 6.5 & \underline{0.73} & 15.1 & 0.75 & 12.7 \\
    \addlinespace[3pt]
    SEVA & \underline{4.3} & \textbf{0.99} & \underline{4.0} & 0.89 & \textbf{2.5} & \textbf{0.95} & \textbf{4.8} & \textbf{0.96} & \underline{3.9} \\
    SV3D & \textbf{3.9} & \textbf{0.99} & \textbf{2.5} & 0.91 & \textbf{2.5} & \textbf{0.95} & \underline{6.0} & \underline{0.95} & \textbf{3.7} \\
    \cmidrule(l){1-10}
    Ours & 7.6 & \underline{0.98} & 5.9 & \underline{0.94} & \underline{3.0} & \textbf{0.95} & 6.9 & 0.93 & 5.9 \\
    \bottomrule
  \end{tabular}
\end{table*}

\section{Text-to-3D Generation Experiments}

\subsection{Baselines} 

We compare our text-to-3D pipeline against SotA text-to-3D methods: GaussianDreamer~\cite{yi2024gaussiandreamer}, DiffSplat (SD3.5m)~\cite{lin2025diffsplat}, TRELLIS-text (xlarge)~\cite{xiang2025structured}, and MVDream~\cite{shi2024mvdream}. All of them output 3D Gaussian splats except MVDream, which outputs a NeRF (marked $^\dagger$). Each baseline is run with the default parameters and with the same seed numbers as our method.
 
Every baseline receives the prompt ``A photo of $\langle$species$\rangle$ plants. Single potted plant, whole plant.'' The first sentence is the guidance prompt of PlantDreamer~\cite{hartley2025plantdreamer}, whose real plants serve as our reference set in Section~\ref{subsec:synthetic}, and the second asks every method for a single, complete potted plant, matching more closely to the reference plants against which all methods are scored. Our method receives the same content prompt followed by the framing suffix described in Section~\ref{subsec:textto3d}. PlantDreamer itself is not included as a baseline, since it requires an L-system mesh as input rather than text alone, and it was trained on the same plants that form our reference set.
 
In the fruit-count experiment (Section~\ref{subsec:synthetic_counting}) we additionally include TRELLIS.2~\cite{xiang2026native}, one of the strongest of the 3D generators evaluated in Section~\ref{sec:img_recons_experiments}, and give it the same Qwen-Image-2.1 input images as our method. Since both methods then start from an identical image, any difference between them in the fruit count is attributable to the 3D lifting stage alone. 

\subsection{Fully synthetic plants}
\label{subsec:synthetic}

We generate bean, kale and mint, the three species of the PlantDreamer dataset (Section~\ref{subsec:datasets}), so that the generated plants can be compared with real ones. Each method generates 12 assets per species, with seeds 1--11 and 42, giving 36 assets per method.
 
\emph{Appearance} is assessed with three image metrics, each averaged over the 72 views of an asset: the CLIP score~\cite{radford2021learning} ($100\times$ the cosine similarity between the image and prompt embeddings of OpenAI CLIP ViT-L/14); ImageReward~\cite{xu2023imagereward}, a learned model of human preference for text-to-image outputs; and the VBench imaging-quality score~\cite{huang2024vbench}, a no-reference measure of distortions such as blur and noise. For every method, CLIP and ImageReward score each view against the bare prompt ``A photo of $\langle$species$\rangle$ plants'', so that neither the object sentence nor our framing suffix affects these scores.
 
For \emph{realism}, we compute the kernel distance (KD) between the rendered views and real photographs of the same species~\cite{binkowski2018demystifying}. The reference set consists of photographs of the 11 bean, 3 kale and 6 mint plants of the PlantDreamer dataset described in Section~\ref{subsec:datasets}, with the background removed in the same way, excluding the photographs used to train PlantDreamer and restricted to camera elevations of $-5^\circ$ to $35^\circ$ to match the renders. We compute KD on Inception-v3 pool features (reported $\times100$) and on DINOv2-B CLS features~\cite{oquab2023dinov2}. 
 
To assess \emph{shape}, we score each asset against PlantDreamer's real point clouds of the same species with the F-score~\cite{tatarchenko2019single}, after scaling both clouds to unit height and aligning them rigidly. As a generated plant is compared with a different real plant rather than with its own reference, we use a looser threshold, $\tau = 2\%$ of plant height, chosen on real plants alone: two real plants of the same species score 0.66 at this threshold, against 0.54 at 1\% and a saturated 0.85 at 5\%. Following~\cite{achlioptas2018learning}, MMD-F averages over the real plants the F-score of the closest generated asset, measuring how closely the best generated shapes match real plants, and COV is the fraction of real plants that are the nearest neighbour of at least one generated asset, measuring how many distinct real shapes are covered. As a reference level, the real plants compared with one another reach an MMD-F of 0.659 and a COV of 0.510.
 
Table~\ref{tab:t23d_phase1} shows that our method achieves the highest CLIP score and ImageReward, above every baseline. Our assets correspond most closely to the requested species and are the most preferred under a learned model of human preference. They are, however, less sharp than TRELLIS-text, MVDream and DiffSplat. A likely cause is that our splats are fitted to a single, nearly level orbit, so the views at the higher evaluation elevations lie outside the generated trajectory, where our reconstructions are prone to artefacts.
 
Realism depends on the feature space. ImageNet-trained Inception features favour texture over shape~\cite{geirhos2018imagenet}, while DINOv2 features capture object structure and track human judgements more closely~\cite{stein2023exposing}. Using DINOv2 features, our model ranks second, behind only TRELLIS-text. Conversely, on Inception features, it ranks last among the five methods. This disparity is mainly driven by the "mint" ($30.54$). Our method generates dense, bushy plants, often in terracotta pots, whereas the reference mint plants are sparse and leggy and sit in black pots.
 
On shape, our method achieves the highest MMD-F, showing our closest generated assets match real plant shapes most accurately. However, its lower coverage indicates our shapes are less diverse across seeds than baseline outputs. 

\begin{figure*}
  \centering
  \includegraphics[width=\textwidth]{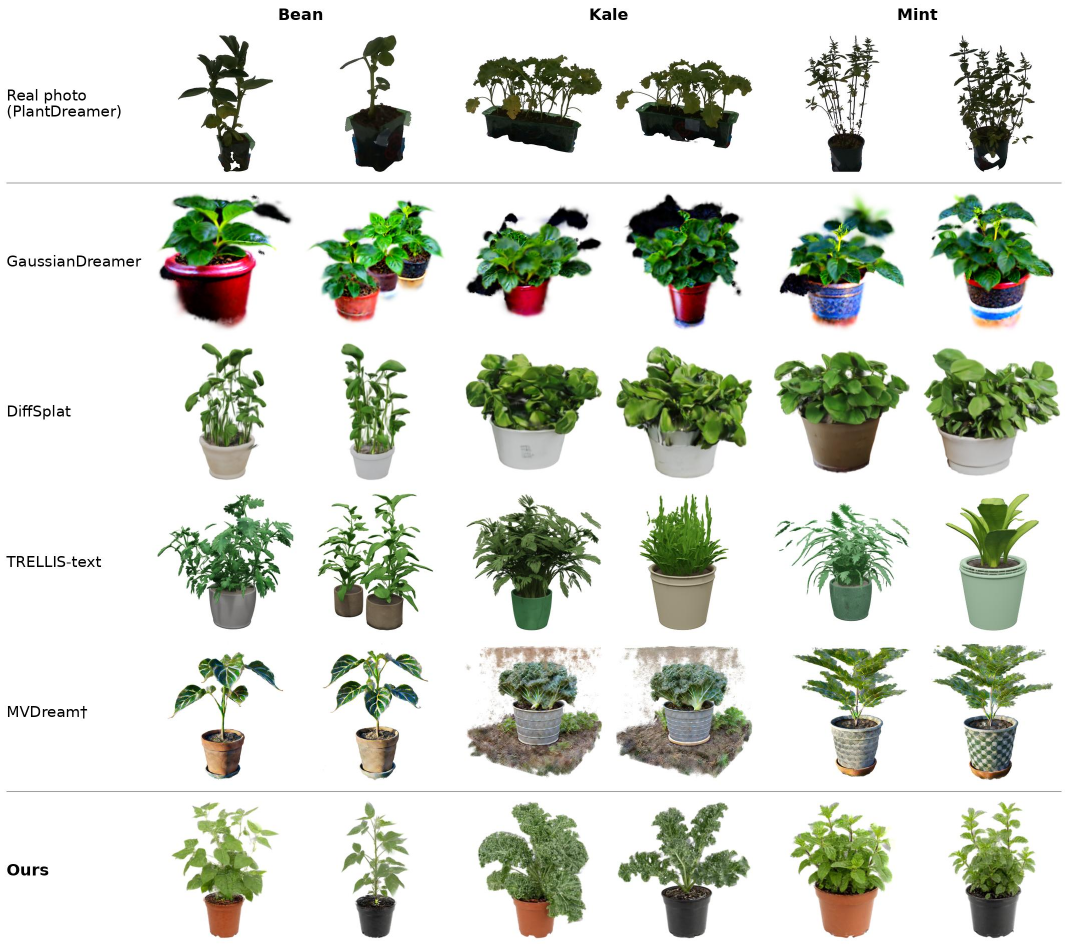}
  \caption{Text-to-3D samples (seeds 1 and 2). Every baseline receives the prompt ``A photo of
    $\langle$species$\rangle$ plants. Single potted plant, whole plant.'' Ours receives the same text with the
    final period replaced by its text-to-image suffix (``\ldots whole plant and pot fully in frame and centred with
    margin around it, eye-level side view, isolated on a plain white background, studio lighting.''), generates an
    image from it and lifts that image to 3D. 
    $^\dagger$NeRF output.}
  \label{fig:t23d_phase1_samples}
\end{figure*}

Figure~\ref{fig:t23d_phase1_samples} shows the generated plant samples per species for each method, together with real PlantDreamer photographs. Compared with these real plants, our assets closely resemble those of the real species, though real mint samples are noticeably taller and sparser than ours, which is consistent with our larger KD on mint. 

In summary, although our method trails the TRELLIS-text slightly in realism, it demonstrates superior prompt adherence in generating specified plant species and achieves the most accurate structural match to real plant geometry.

\begin{table*}
  \caption{Text-to-3D plant generation: 12 seeds per species (bean, kale, mint), $n{=}36$ assets per method. Appearance: mean over 72 views per asset. Realism: kernel distance to PlantDreamer's
    real photos, mean over species, 95\% bootstrap interval in brackets; Shape: against PlantDreamer's real point clouds; real plants vs each other: MMD-F
    0.659, COV 0.510. \textbf{Bold} / \underline{underline}: best / second-best text-only method. $^\dagger$NeRF output.}
  \label{tab:t23d_phase1}
  \centering
  \footnotesize
  \setlength{\tabcolsep}{4pt}
 \begin{tabular}{@{}lccc@{\hspace{8pt}}cc@{\hspace{8pt}}cc@{}}
    \toprule
     & \multicolumn{3}{c}{Appearance} & \multicolumn{2}{c}{Realism} & \multicolumn{2}{c}{Shape} \\
    \cmidrule(lr){2-4} \cmidrule(lr){5-6} \cmidrule(l){7-8}
    Method & CLIP$\uparrow$ & ImageReward$\uparrow$ & Imaging$\uparrow$
      & KD$_\text{Incep}$$\downarrow$ & KD$_\text{DINOv2}$$\downarrow$ & MMD-F$\uparrow$ & COV$\uparrow$ \\
    \midrule
    GaussianDreamer       & \underline{23.41} & \underline{$-$0.935} & 0.382 & \underline{22.74\,{\scriptsize[20.8,\,25.0]}} & 53.44\,{\scriptsize[51.5,\,55.4]} & 0.432 & 0.308 \\
    DiffSplat             & 20.54 & $-$1.340 & 0.457 & 24.83\,{\scriptsize[22.8,\,27.3]} & 58.44\,{\scriptsize[56.6,\,60.2]} & 0.480 & \textbf{0.475} \\
    TRELLIS-text          & 19.89 & $-$1.806 & \textbf{0.600} & \textbf{18.20\,{\scriptsize[15.8,\,21.1]}} & \textbf{35.26\,{\scriptsize[32.8,\,37.9]}} & \underline{0.514} & \underline{0.419} \\
    MVDream$^\dagger$     & 20.55 & $-$1.349 & \underline{0.542} & 23.99\,{\scriptsize[22.1,\,26.2]} & 60.97\,{\scriptsize[59.5,\,62.5]} & 0.491 & 0.338 \\
    \cmidrule(l){1-8}
    Ours                  & \textbf{24.99} & \textbf{$-$0.687} & 0.412 & 25.30\,{\scriptsize[22.8,\,28.1]} & \underline{39.46\,{\scriptsize[36.9,\,41.8]}} & \textbf{0.539} & 0.288 \\
    \bottomrule
  \end{tabular}
\end{table*}

\subsection{Synthetic Fruit Counting Experiments}
\label{subsec:synthetic_counting}
To test whether a fine-grained attribute specified in the prompt is preserved in 3D, we prompt each method for a given number of fruit, using ``A photo of $\langle$species$\rangle$ plants with $\langle N\rangle$ $\langle$fruit$\rangle$. Single potted plant, whole plant.'' (e.g.\ ``with five tomatoes''). We write $N$ as a word, since text-to-image models follow number words more accurately than digits~\cite{kajic2024evaluating}, which we also observed for Qwen-Image-2.1 in a pilot study. We generate tomato and lemon plants for $N=1$ to $7$ with seeds 42, 1 and 2, giving 42 assets per method. As in Section~\ref{subsec:synthetic}, our method receives the same content prompt followed by the framing suffix.

Fruit are counted by eye by one author, blind to the requested number and method, from rendered views covering each asset from all sides. Each distinct fruit of any colour is counted once, including fused, detached and blurred fruit, but not colour on a leaf or stem and specks visible from only one angle. A second author, independently re-counted one asset per method, species and $N$ (84 of the 252 assets) and 14 of the 42 input images. The two counters agreed exactly on 91.7\% of the assets (Krippendorff’s $\alpha=0.98$~\cite{krippendorff2018content}) and on all 14 input images. Every disagreement concerned the ill-formed fruit of DiffSplat (5 of 14 assets) or MVDream (2 of 14), yet none changed whether an asset matched the requested number, so each method’s exact-count rate on the subset is the same under either counter. Table~\ref{tab:t23d_counting} uses the reported counts of the first counter.

We report the percentage of the percentage of assets whose counted fruit equals the requested $N$ (Exact Number Generation), the mean absolute count error (MAE) and, for our method and TRELLIS.2, the percentage whose 3D count matches their input image ($=$input). CLIP, ImageReward and VQAScore~\cite{lin2024evaluating}, which uses a vision-language model to estimate the probability that an image shows the prompt, are averaged over the 72 views and computed without the count (e.g.\ ``A photo of tomato plants with tomatoes’’), so that they measure image quality and the presence of fruit.

\begin{figure*}
  \centering
  \includegraphics[width=\textwidth]{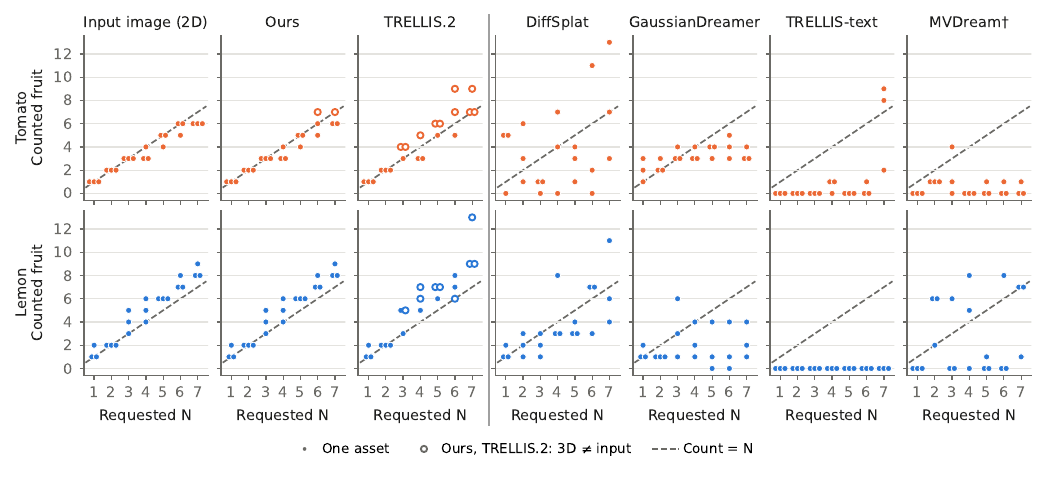}
  \caption{Counted fruit against the requested number $N$ for every asset.
    Each dot is one asset; Rows: tomato and
    lemon, 21 assets each per method ($N = 1$--$7$, seeds 42, 1 and 2). Dots on the dashed line have exactly $N$ fruit; \textit{Input image}: the count in the input image, which was
    then used as input to ours and TRELLIS.2. Hollow dots: the 3D asset's count differs from its input image's (ours: 2 of 42, both one more;
    TRELLIS.2: 19 of 42, 18 more and 1 fewer). $^\dagger$NeRF output.}
  \label{fig:t23d_counting_scatter}
\end{figure*}

Our method preserves the fruit count of its input image in 40 of the 42 assets (95.2\%; Table~\ref{tab:t23d_counting}).The two exceptions are both tomato assets whose input image shows six fruit and whose 3D asset gains one, giving seven (see Fig. \ref{fig:t23d_counting_scatter}). For the $N=6$ asset, this turns a correct count into an error, while for the $N=7$ asset it turns an error into a correct count. The two changes cancel out, so our Exact Number Generation and MAE are identical to those of the input images. TRELLIS.2, lifting the same images, preserves the count in only 23 of 42 assets (54.8\%), and 18 of its 19 errors add fruit.  We attribute this difference to how each method uses the input image. In our pipeline, the input image is pinned as the first and last frame of the orbit video, which only moves the camera around a static scene (Section~\ref{subsec:orbitvideo}), so the visible fruit are carried into the splat. TRELLIS.2 instead uses the image only as conditioning and generates the whole asset, including its hidden side, from its 3D prior, leaving it free to add fruit.

Because our 3D counts follow the input images so closely, our remaining errors originate in the text-to-image stage. As Fig.~\ref{fig:t23d_counting_scatter} shows, these errors depend on the species, with tomato images having at most one fruit too few and lemon images up to two too many, and they become more frequent as $N$ grows.

End-to-end, our method achieves the highest Exact Number Generation and the lowest MAE of all methods. Notably, TRELLIS-text, which produced the most realistic renders in Section~\ref{subsec:synthetic}, never produces the requested count (0.0\%, MAE 3.62), as it mostly omits the fruit altogether (Fig.~\ref{fig:t23d_counting_scatter}). On the quality metrics, our method achieves the highest VQAScore, close to that of its input images, and ranks second on CLIP and ImageReward, behind TRELLIS.2. These differences are small, and as TRELLIS.2 is rendered as a textured mesh and our method as a splat, they may partly reflect the different output representations.

\begin{figure*}
  \centering
  \includegraphics[width=\textwidth]{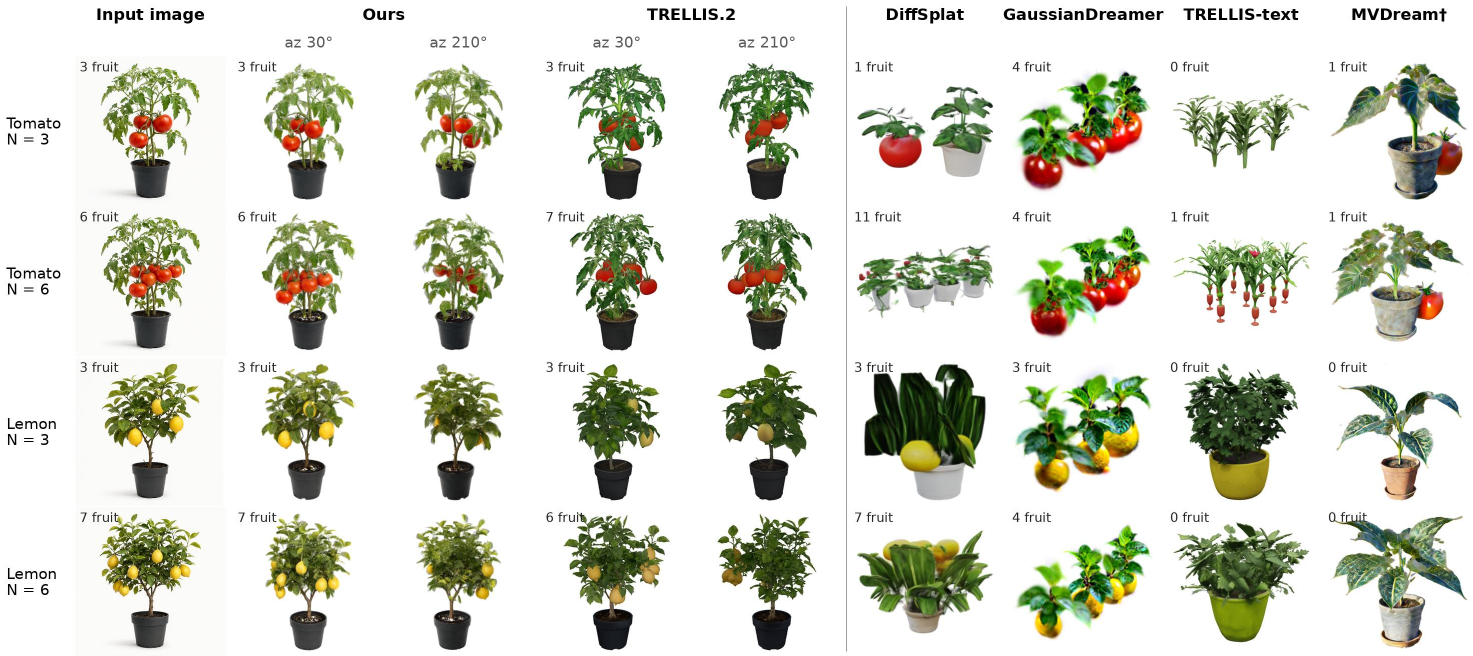}
  \caption{Text-to-3D samples for prompted fruit counts: tomato and lemon with
    $N = 3$ and $N = 6$. Every baseline
    receives ``A photo of $\langle$species$\rangle$ plants with $\langle N\rangle$ $\langle$fruit$\rangle$. Single
    potted plant, whole plant.'' with the number written as a word;  The input image (left) is generated by Qwen-Image-2.1~\cite{qwen2026qwenimage21}
    from the same text, with the final period replaced by the text-to-image suffix (``\ldots whole plant and pot fully
    in frame and centred with margin around it, eye-level side view, isolated on a plain white background, studio
    lighting.''); ours and TRELLIS.2 both lift this image to 3D. Two views of ours and
    TRELLIS.2 from opposite sides (elevation $15^\circ$, azimuth $30^\circ$ and $210^\circ$), one ($30^\circ$) of each text-to-3D baseline. $^\dagger$NeRF output.}
  \label{fig:t23d_counting_samples}
\end{figure*}

Figure~\ref{fig:t23d_counting_samples} shows examples for $N=3$ and $N=6$. The text-to-3D baselines often generate malformed or missing fruit, and mostly match the requested count at a rate close to or below random chance (14.3\%). These failures are consistent with the observation in Section~\ref{sec:related work} that text-to-3D models, trained on far less data than text-to-image models, offer little control over specific plant traits. Our method and TRELLIS.2 avoid this limitation by starting from an image generated by a strong text-to-image model.

Overall, our method follows the requested fruit count more accurately than every text-to-3D baseline and almost always preserves the count of its input image, so its accuracy is limited mainly by the text-to-image model and should improve with it.

\begin{table*}
  \caption{Prompted fruit counts (requested $N = 1$--$7$; $3$ different seeds per species (tomato, lemon); $n{=}42$ assets per method). Fruit counted by eye over each asset's 36 evaluation views and turntable, each fruit once, all assets included; agreement with a blinded second counter on a subset: Krippendorff's $\alpha = 0.98$. Exact Number Generation: count $= N$ (Random Chance, uniform guess over 1–7: $14.3\%$). MAE: mean absolute count error. $=$input: the 3D count equals the count in the input image. \textit{Input image}: the 2D image that ours and TRELLIS.2 start from. CLIP, ImageReward and VQAScore use the prompt without the count.
    \textbf{Bold} / \underline{underline}: best / second-best 3D method. $^\dagger$NeRF output. $^\ddagger$mesh output.}

  \label{tab:t23d_counting}
  \centering
  \footnotesize
  \setlength{\tabcolsep}{4pt}
  \begin{tabular}{@{}lccc@{\hspace{8pt}}ccc@{}}
    \toprule
    Method & Exact Number Generation (\%)$\uparrow$ & MAE$\downarrow$ & $=$input (\%)$\uparrow$
           & CLIP$\uparrow$ & ImageReward$\uparrow$ & VQAScore$\uparrow$ \\
    \midrule
    GaussianDreamer            & 23.8 & 1.88 & -- & 25.37 & $-$0.26 & \underline{0.765} \\
    DiffSplat                  & 14.3 & 2.17 & -- & 23.98 & $-$1.57 & 0.483 \\
     TRELLIS-text               & 0.0 & 3.62 & -- & 19.93 & $-$2.10 & 0.128 \\
    MVDream$^\dagger$          & 7.1 & 3.29 & -- & 21.82 & $-$1.65 & 0.336 \\
    \cmidrule(l){1-7}
    TRELLIS.2$^\ddagger$       & \underline{40.5} & \underline{1.02} & 54.8 & \textbf{28.35} & \textbf{$-$0.02} & 0.607 \\
    Ours                       & \textbf{50.0} & \textbf{0.60} & \textbf{95.2} & \underline{27.58} & \underline{$-$0.13} & \textbf{0.867} \\
    \midrule
    \textit{Input image (2D)}  & 50.0 & 0.60 & -- & 29.14 & 0.47 & 0.913 \\

    \bottomrule
  \end{tabular}
\end{table*}

\section{Discussion}
\label{sec:discussion}

\subsection{Advantages over existing methods}
\label{advantages}
We believe that Plant Orbiter presents several advantages over existing approaches to 3D reconstruction of plants. Firstly, monocular methods of 3D reconstruction present numerous practical advantages for phenotyping applications. High throughput pipelines, like those normally used in plant science and agriculture, must work efficiently and at scale, making capture using complex imaging setups or expensive robots impractical. Efficient, one or few shot reconstruction approaches, captured with a handheld camera present a much more practical approach in a wide range of settings.

Relative to other generation methods such as TRELLIS or Hunyuan3D, our work offers a substantial practical advantage by being able to easy align the model to its input image; even with respect to other video diffusion models, the ability to predict camera poses based on the known path provides computational benefits at scale.

When compared to other reconstruction from video, including real and synthetic methods like SEVA or SV3D, which inherit part of our pipeline, our method beats both methods across a majority of the evaluations conducted. Our strongest results against these methods come from Shape and Perceptual measures, for which we beat SEVA in all tests, and SV3D in all but one. All three video methods use our known poses and therefore skip the lengthy COLMAP process, a stage that can often lead to reconstruction failure. For complex plant scenes it is often challenging to reconstruct camera poses using COLMAP, and the inclusion of our known pose approach greatly strengthens this family of methods.

For text-to-3D generation, our results suggest that the priors of 2D and video models can stand in for scarce 3D training data. The text-to-image model provides control over the species and traits of the plant, such as the number of fruit, and the video model, by orbiting the input image rather than regenerating the plant, carries that control into 3D. The prompt therefore remains an accurate label for the generated plant, as labelled synthetic data for phenotyping requires, and the pipeline should benefit directly from future text-to-image models.

For both reconstruction and generative pipelines we present, our method is especially of interest as it achieves these results without any plant specific training. Many generative~\cite{hartley2025plantdreamer} models for plants require training from plant specific datasets that are often species and setup dependant, meaning that these models do not generalise well to unseen cases. By contrast, our method performs well across a wide variety of plant species without requiring additional training.

\subsection{Limitations}
\label{limiations}
The biggest limitation of this work is that in phenotyping pipelines, monocular approaches can only capture details included in the input image, and details of the reverse side, or occluded features must be inferred. For this reason our method is best suited to specific use cases, like height measurement as shown in the paper, and specific plants that do not self occlude significantly. 

Another limitation of our method is that the current path predicted captures a single orbit of the plant at a static height. Since Gaussian splatting is trained from this video from the specific orbital perspective, the reconstructions often have artefacts and loss of realism, when viewed from above and below. Though the impact of this limitation will vary in importance depending on the specific downstream phenotyping task being attempted.

\subsection{Future Work}
\label{future_work}
One of the primary limitations of any monocular approach is that the rear view of the plant will, by the nature of this approach, be invented. As such, a video diffusion model that takes an additional image of the reverse view of the plant may be required for further phenotyping tasks, or dense plant structures where additional knowledge of the far side is required. An adapted version of our model that pins an additional view could allow for this, but would need to overcome additional challenges in plant consistency, and the ability to determine the placement of the camera in relation to each other, and track the camera path between them.

Additional work on synthetic data generated this way is also highly promising. Compared to other methods of generating synthetic plant data, for example, procedural generative methods, a crucial advantage of text-to-3D generative models is the ability to prompt for specific phenotypic characteristics easily. Further investigation into the level of accuracy (in the case of geometric measures) or realism when generating physiological phenotypes that can be achieved with our method could strengthen our methods potential for synthetic data generation.

\subsection{Conclusion}
\label{Conclusion}
In this paper we have presented Plant Orbiter, to our knowledge the first monocular 3D reconstruction method for plants using video diffusion. Our experimental results show that our method competes strongly against existing SotA methods, often exceeding their performance in crucial metrics such as 3D geometry across varied datasets.

Our method generates a short 73 frame video around a target plant, from which we generate a 3D Gaussian splat, using the known orbit from around the plant to predict accurate camera positions and easily align the generated 3D plant. When tested on plants from a wide variety of species, we find we are able to achieve the highest accuracy in IoU and DISTS for most datasets and the best F-scores on all tested datasets. We demonstrate how this method can be used to measure real world plant traits such as plant height and achieve a mean height error of just 5.9\% (0.9-2.3cm) across our experiments. 

By prepending a text-to-image model, the same pipeline also generates 3D plants from text, inheriting the prompt adherence of 2D models without requiring scarce 3D training data. Among the text-to-3D methods tested, it best matches both the requested species and the shapes of real plants, reproduces a desired number of fruit accurately, and preserves the fruit count of its input image in 95\% of assets.

Overall, these results show an accurate and realistic new way to reconstruct digital twins, perform 3D phenotyping, and generate synthetic data using a method not yet applied in this field, while also providing meaningful practical benefits to users.

\section*{Declaration of generative AI and AI-assisted technologies in the manuscript preparation process}
During the preparation of this work the authors used Claude (Anthropic) in order to assist with coding, organising and documenting the code repository, and figure improvement. After using this tool, the authors reviewed and edited the content as needed and take full responsibility for the content of the publication.

\section*{Data Availability}
All datasets used in this work are from previous publications and are publicly available. All code as well as orbit videos and reconstructions will be available from the corresponding author upon reasonable request.

\section*{Funding}
Yong En Kok is supported by the Biotechnology and Biological Sciences Research Council [UKRI3716] and Lewis Stuart is supported by the Engineering and Physical Sciences Research Council [EP/W524402/1].

\section*{CRediT authorship contribution statement}
\textbf{Zane K.J. Hartley} Conceptualisation, Methodology, Software, Validation, Formal Analysis, Investigation, Data Curation, Visualisation, Writing - original draft, Writing - review and editing. \textbf{Yong En Kok} Conceptualisation, Methodology, Software, Validation, Formal Analysis, Investigation, Data Curation, Visualisation, Writing - original draft, Writing - review and editing. \textbf{Lewis A.G. Stuart} Conceptualisation, Writing - review and editing \textbf{Andrew P. French} Supervision, Writing - review and editing \textbf{Michael P. Pound} Supervision, Writing - review and editing

\section*{Declaration of competing interest}
The authors of this paper have no competing interests to declare.

\bibliographystyle{cas-model2-names}

\bibliography{cas-refs}

\end{document}